\documentclass{sig-alternate-05-2015}

\usepackage{todonotes}

\usepackage{times}
\usepackage{url}
\usepackage{latexsym}
\usepackage{cite}
\usepackage{amsmath}
\usepackage{booktabs}
\usepackage{amssymb}
\usepackage{graphicx}
\usepackage{caption}
\usepackage{multirow}
\usepackage{subfig}
\usepackage{dsfont}

\usepackage{xcolor,colortbl}
\usepackage{array}
\newcolumntype{L}[1]{>{\raggedright\let\newline\\\arraybackslash\hspace{0pt}}m{#1}}
\newcolumntype{C}[1]{>{\centering\let\newline\\\arraybackslash\hspace{0pt}}m{#1}}
\newcolumntype{R}[1]{>{\raggedleft\let\newline\\\arraybackslash\hspace{0pt}}m{#1}}

\providecommand{\e}[1]{\ensuremath{\times 10^{#1}}}

\begin{document}

\setcopyright{rightsretained}

\doi{}

\isbn{}

\conferenceinfo{Neu-IR 16 SIGIR Workshop on Neural Information Retrieval}{July 21, 2016, Pisa, Italy}

\acmPrice{}

%

\title{Uncertainty in Neural Network Word Embedding \\ Exploration of Threshold for Similarity}

%
%
%
%
%

%
\author{
%
%
\alignauthor
Navid Rekabsaz, Mihai Lupu, Allan Hanbury\titlenote{This work is partly funded by two projects: SelfOptimizer (9867) by EuroStar and  ADMIRE (P 25905-N23) by FWF. Thanks to Joni Sayeler and Linus Wretblad for their contributions in the SelfOptimizer project.}\\
       \affaddr{Vienna University of Technology}\\
       \affaddr{Favorittenstrasse 9}\\
       \affaddr{Vienna, Austria}\\
       \email{$family\_name$@ifs.tuwien.ac.at}
}

\maketitle
\begin{abstract}
Word embedding, specially with its recent developments, promises a quantification of the similarity between terms. However, it is not clear to which extent this similarity value can be genuinely meaningful and useful for subsequent tasks. We explore how the similarity score obtained from the models is really indicative of term relatedness. We first observe and quantify the uncertainty factor of the word embedding models regarding to the similarity value. Based on this factor, we introduce a general threshold on various dimensions which effectively filters the highly related terms. Our evaluation on four information retrieval collections supports the effectiveness of our approach as the results of the introduced threshold are significantly better than the baseline while being equal to  or statistically indistinguishable from the optimal results. 
\end{abstract}

\vspace{-0.3cm}
\section{Introduction}
\label{sec:introduction}
Understanding the meaning of a word (semantics) and of its similarity to the other words (relatedness) is the core of understanding text. An established method for quantifying this similarity is the use of \emph{word embeddings}, where vectors are proxies of the meaning of words and distance functions are proxies of semantic and syntactic relatedness. Fundamentally, word embedding models exploit the contextual information of the target words to approximate their meaning, and hence their relations to the other words.


Given the vectors representing words and a corresponding mathematical function, word embedding models  provide an approximation of the the relatedness of any two terms, although this relatedness could be perceived as completely-meaningless in the language. An emerging challenge here is: \emph{how to identify whether the similarity score obtained from word embedding is really indicative of term relatedness?}. This issue is pointed out by  Karlgren et al.~\shortcite{karlgren2008} in examples, showing that word embedding methods are too ready to provide answers to meaningless questions: \emph{``What is more similar to a computer: a sparrow or a star?''}, or \emph{``Is a cell more similar to a phone than a bird is to a compiler?''}. 

In the absence of a comprehensive answer, the need for related terms has been generally met by applying $k$ Nearest Neighbours ($k$-NN) search such that retrieving the top $k$ most similar terms in the neighbouring of a given term as related terms. Recently, Cuba Gyllensten and Sahlgren~\shortcite{gyllensten2015navigating} point out the limitations of the $k$-NN approach as it neglects the internal structure of neighbourhoods which could be vastly different for various terms. In other words, some terms are more central in language and therefore have more related terms while many words have no genuinely related term. This is intuitive in human language while also quantifiable by using a language thesaurus e.g. WordNet (for example, by counting the number of synonyms). We therefore put the focus of this study on the notion of  ``similar'' in word embedding.

Different characteristics of  term similarities have been explored in several studies: the concept of relatedness~\cite{kruszewski2015so,kiela2015specializing}, the similarity measures~\cite{koopman2012evaluation}, intrinsic/extrinsic evaluation of the models~\cite{schnabel2015evaluation,Tsvetkov2015EvaluationOW,baroni2014don,de2014medical}, or in sense induction task~\cite{gyllensten2015navigating,erk2010exemplar}. However, there is lack of understanding on the internal structure of word embedding, specifically how its similarity distribution reflects the relatedness of terms.   

Following this direction, in this work, we would argue that the ``similar'' words can be identified by a threshold on similarity values which separates the semantically related words from the less or non-related ones. It is quite difficult, a priori, to even consider a threshold for this similarity. Especially since we do not want to make this parameter dependent on the term. This would be not only computationally, but also conceptually problematic. As Karlgren et al. discuss for the case of Random Indexing~\shortcite{karlgren2008,karlgren2014}, just because we \emph{can} have a ``most similar term(s)'' does not mean that this makes any sense in real life. 

Certainly, the meaning of ``similar'' also depends on the similarity function, but, we consider here the state of the art word similarity and leave the exploration of this factor for the further studies. Instead, we would argue that regardless of the similarity function, the most important factor is the threshold which separates the semantically related terms from the less or non-related ones.

Exploring such a threshold has the potential to bring improvements in those studies which use word embedding for retrieving the similar/related words in different tasks i.e. query expansion~\cite{grbovic2015context}, query auto-completion~\cite{mitra2015exploring}, document retrieval ~\cite{rekabsaz2015content}, learning to rank~\cite{severyn2015learning}, language modelling in IR~\cite{ganguly2015word}, or Cross-Lingual IR~\cite{vulic2015monolingual}.

We explore the estimation of this potential threshold by first quantifying the \emph{uncertainty} factor in the similarity values of embedding models.  This factor is an intrinsic characteristic of all the recent models, because they all  start with some random initialization and eventually converge to a (local) solution. Therefore, even by training with the same parameters and on the same data, the created word embedding models result in slightly different word distributions and hence slightly different relatedness values. In the next step, using the \emph{uncertainty} factor, we provide a continuous neighbouring representation for an arbitrary term, which is later used to estimate the general threshold.   

In order to evaluate the effectiveness of the introduced threshold, we test it in the context of a document retrieval task, on five different test collections. In the experiments, we apply the  threshold to identify the set of terms to meaningfully extend the query terms. We show that using the introduced threshold performs either exactly the same as or statistically indistinguishable from the optimal threshold for all collections.

In summary, the main contributions of the current study are:
\begin{enumerate}
\vspace{-0.2cm}\item exploration of the uncertainty factor in word embedding models in different dimensions and similarity ranges.
\vspace{-0.2cm}\item introducing a general threshold for separating similar terms in different dimensions.
\vspace{-0.2cm}\item extensive experiments on five test collections comparing different threshold values as well as $k$-NN search. 
\end{enumerate}

Among various word embedding models, in our study, we use the method proposed by Mikolov et al.~\cite{mikolov2013efficient}: skip-gram with negative-sampling training (SGNS) method in the  Word2Vec framework. While this is not the newest method in this category (e.g. Pennington et al.~\shortcite{pennington2014glove} introduced GloVe and reported superior results), independent benchmarking provided by Levy et al.~\shortcite{levy2015improving} shows that there is no fundamental performance difference between the recent word embedding models. In fact, based on their experiments, they conclude that the performance gain observed by one model or another is mainly due to the setting of the hyper-parameters of the models. Their study also motivates our decision to use SGNS: \textit{``SGNS is a robust baseline. While it might not be the best method for every task, it does not significantly underperform in any scenario.''}  

The remainder of this work is structured as follows: First, we review related work in Section~\ref{sec:related}. We introduce the potential threshold in Section~\ref{sec:analysis}. We present our experimental setup in Section~\ref{sec:experimentsetup}, followed by discussing the results in Section~\ref{sec:evaluation}. Section~\ref{sec:conclusion} summarises our observations and concludes the paper.

\vspace{-0.3cm}
\section{Related Work}
\label{sec:related}
The closest study to our work is Karlgren et al.~\shortcite{karlgren2014}, which explores the semantic topology of the vector space generated by Random Indexing. Based on their previous observations that the dimensionality of the semantic space appears different for different terms~\cite{karlgren2008}, Karlgren at al. now identify the different dimensionalities at different angles (i.e. distances) for a set of specific terms. It is however difficult to map these observations to specific criteria or guidelines for either future models or retrieval tasks.   

In fact, our observations provide a quantification on Karlgren's claim that \emph{``\emph{`close'} is interesting and \emph{`distant'} is not''}~\cite{karlgren2008}. 

More recently, Cuba Gyllensten and Sahlgren~\shortcite{gyllensten2015navigating} follow a data mining approach to represent the terms relatedness by a tree structure. While they suggest traversing the tree as a potential approach, they evaluate it only on the word sense induction tasks and its utility for retrieving similar words remains unanswered. Our work complements and extends  their approach. Defining the  threshold on the collection and not each word, our method is efficiently applicable and computationally cheaper on all subsequent tasks to which the word embeddings may be applied.

\vspace{-0.3cm}
\section{Potential Threshold}
\label{sec:analysis}
As mentioned in introduction, we are looking for a potential threshold to separate the truly related terms from the rest. In this section, we describe our analytic approach to explore such cutting points in different dimensions. The introduced threshold is defined on  the entire model i.e. it is applicable to any arbitrary term in language.

For this purpose, we start with an observation on the uncertainty of similarity in word embedding models, followed by defining a  continuous model of neighbouring distribution, before we define our proposed threshold.
\begin{figure*}
  \centering
  \subfloat[]{\includegraphics[width=0.32\textwidth]{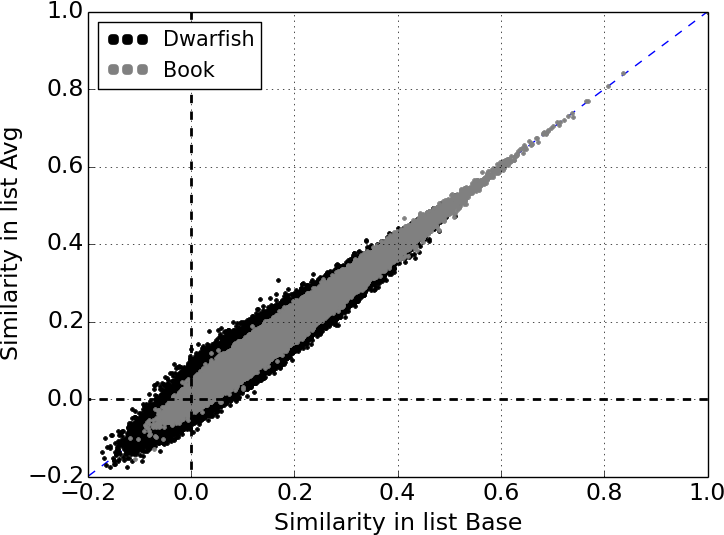}\label{figure:pairs_200}}
  \hfill
\centering
  \subfloat[]{\includegraphics[width=0.32\textwidth]{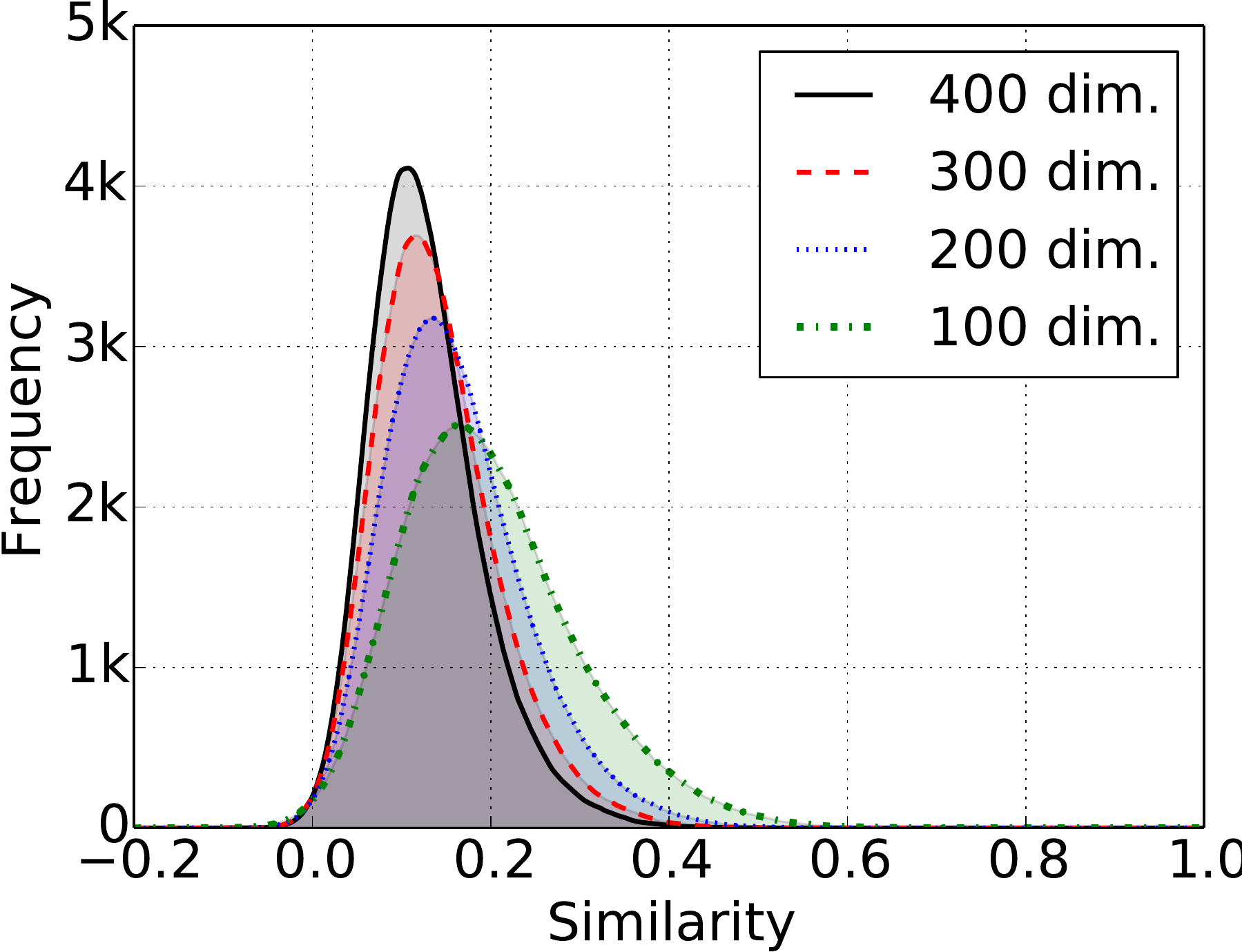}\label{figure:distribution_all}} 
  \hfill
\centering
  \subfloat[]{\includegraphics[width=0.32\textwidth]{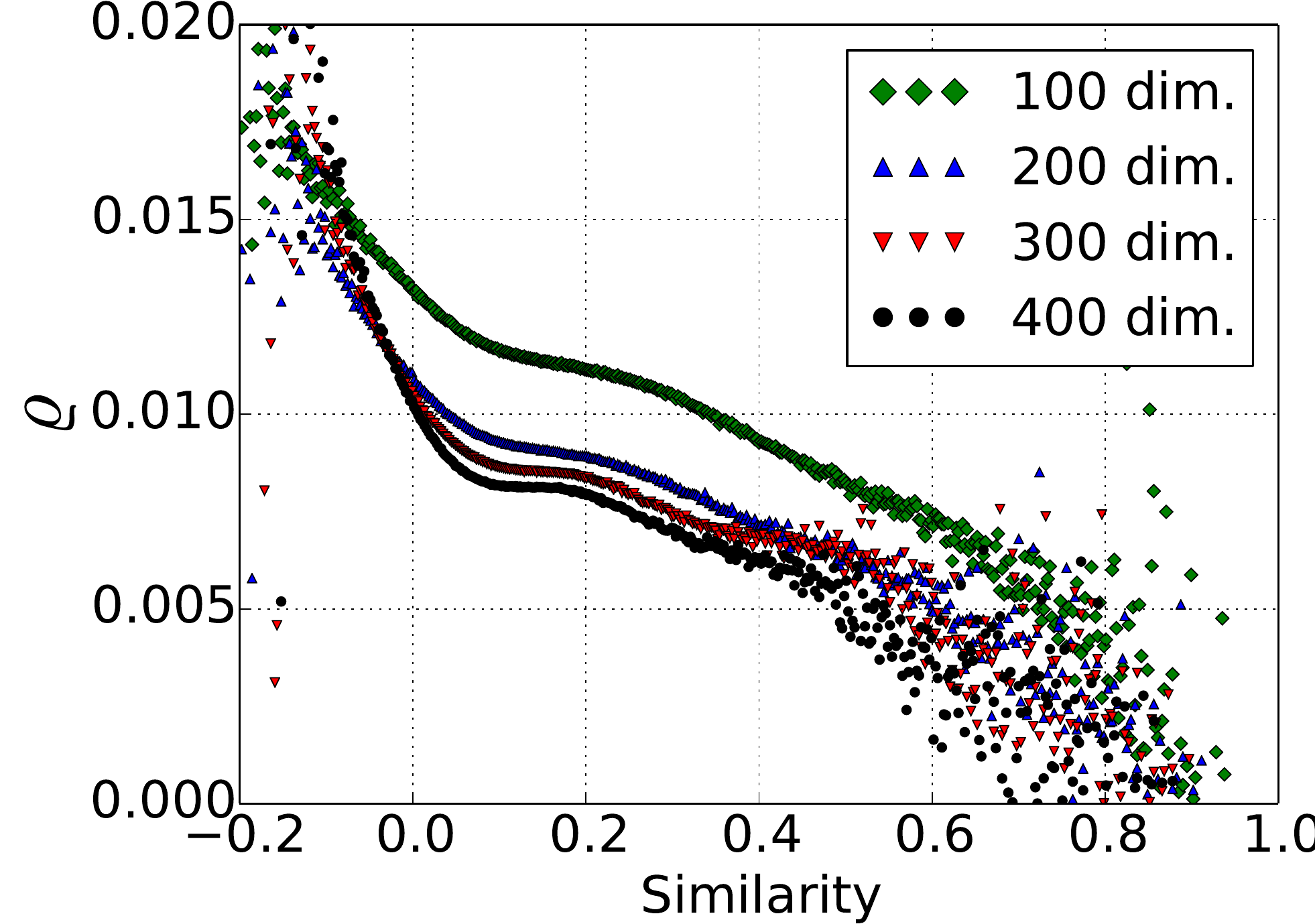}\label{figure:uncertainty_mean}}
  \hfill
  \vspace{-0.4cm}
  \caption{(a) Comparison of similarity values of the terms \emph{Book} and \emph{Dwarfish} to ~580K words between models $M$ and $P$. (b) Histogram of similarity values of an arbitrary term to all the other words in the collection for 100, 200, 300, and 400 dimensions. (c) Average uncertainty for each similarity value span.}
  \label{figure:uncertainty}
  \vspace{-0.4cm}
\end{figure*}
\subsection{Uncertainty of Similarity}\label{sec:sec31}
In this section we make a series of practical observations on word embeddings and the similarities computed based on them.

To observe the uncertainty, let us consider two models $P$ and $M$. 
To create each instance, we trained the Word2Vec SGNS model with the sub-sampling parameter set to $10^{-5}$, context windows of 5 words, epochs of 25, and word count threshold 20 on the Wikipedia dump file for August 2015, after applying Porter stemmer. Each model has a vocabulary of approximately 580k terms. They are identical in all ways except their random starting point.

Figure~\ref{figure:pairs_200} shows the distances between two terms and all other terms in the dictionary, for the two models, in this case of dimensionality 200. For each term we have approximately 580k points on the plot. As we can see, the difference between similarities calculated in the two models, appears (1) greater for low similarities, and (2) greater for a rare word (Dwarfish) than for a common word (Book). We can also observe that there are very few pairs of words with very high similarities.

Let us now explore the effect of dimensionality on similarity values and also uncertainty. Before then, in order to generalize the observations to an arbitrary term, we had to consider a set of ``representative'' terms. What exactly ``representative'' means is of course debatable. We took 100 terms recently introduced in the query inventory method by Schnabel et al.~\shortcite{schnabel2015evaluation}. It is claimed that the terms are diverse in frequency, part of speech (POS). In the following of the paper, we refer to \emph{arbitrary} term as an aggregation over the representative terms. 

Figure~\ref{figure:distribution_all} shows frequency histograms for the occurrence of similarity values in different dimensionalities of a given model. As we can see, similarities are in the $[-0.2,1.0]$ range and have positive skewness (the right tail is longer). As the dimensionality increases, the kurtosis also increases (the histogram has thinner tails).

To observe the changes in uncertainty in different dimensions, we quantify this uncertainty as a function of the similarity value. Let us consider 
\[
\mathcal{S}_s=\left\{(x,y) : sim(\vec{x}_M,\vec{y}_M)\in(s,s+\epsilon) \right\}
\]
the set of term pairs whose similarity is approximately $s$ according to model $M$ ($\vec{x}_M$ is the vector representation of term $x$ in model $M$ and  $sim$ is a similarity function between two vectors (Cosine throughout this paper)). We have to consider this approximation as it is practically never the case that two word pairs have exactly the same similarity value.  We can then define an uncertainty $\varrho$ as follows:
\begin{equation}
  \varrho(s) = \frac{1}{|\mathcal{S}_s|}\sum_{(x,y)\in\mathcal{S}_s}{\left| sim(\vec{x}_M,\vec{y}_M) - sim(\vec{x}_P,\vec{y}_P) \right|}
  \label{formula:rand_wordpair}
\end{equation}
where $\vec{x}_P$ is the vector representation of term $x$ in model $P$. 
The approximation parameter $\epsilon$ is not important for this exemplification. For the plot in Figure~\ref{figure:uncertainty_mean} we take it to be 2.4\e{-4}, as it splits our domain (-0.2,1.0) into 500 equal intervals. 
Figure~\ref{figure:uncertainty_mean} shows  $\varrho$ for different dimensionalities, against the similarity calculated in the $M$ model. We  observe that, as the similarity increases, the uncertainty decreases and that for highly similar words the different model instances tend to agree.


We also observe a decrease in $\varrho$ as the dimensionality of the model increases. On the other hand, the differences between models decrease as the dimension increases such that the models of dimension 300 and 400 seem very similar in comparison to 100 and 200. The observation shows a probable convergence in the Uncertainty at higher dimensionalities.

We can conclude from the observations that the similarity between terms is not a concrete value but can be considered as an approximation whose variation is highly dependent on the dimensionality and similarity range. We use the effect of this factor in the following.

\subsection{Continuous Distribution of Neighbours}\label{sec:sec32}
\begin{figure*}
  \centering
  \subfloat[]{\includegraphics[width=0.32\textwidth]{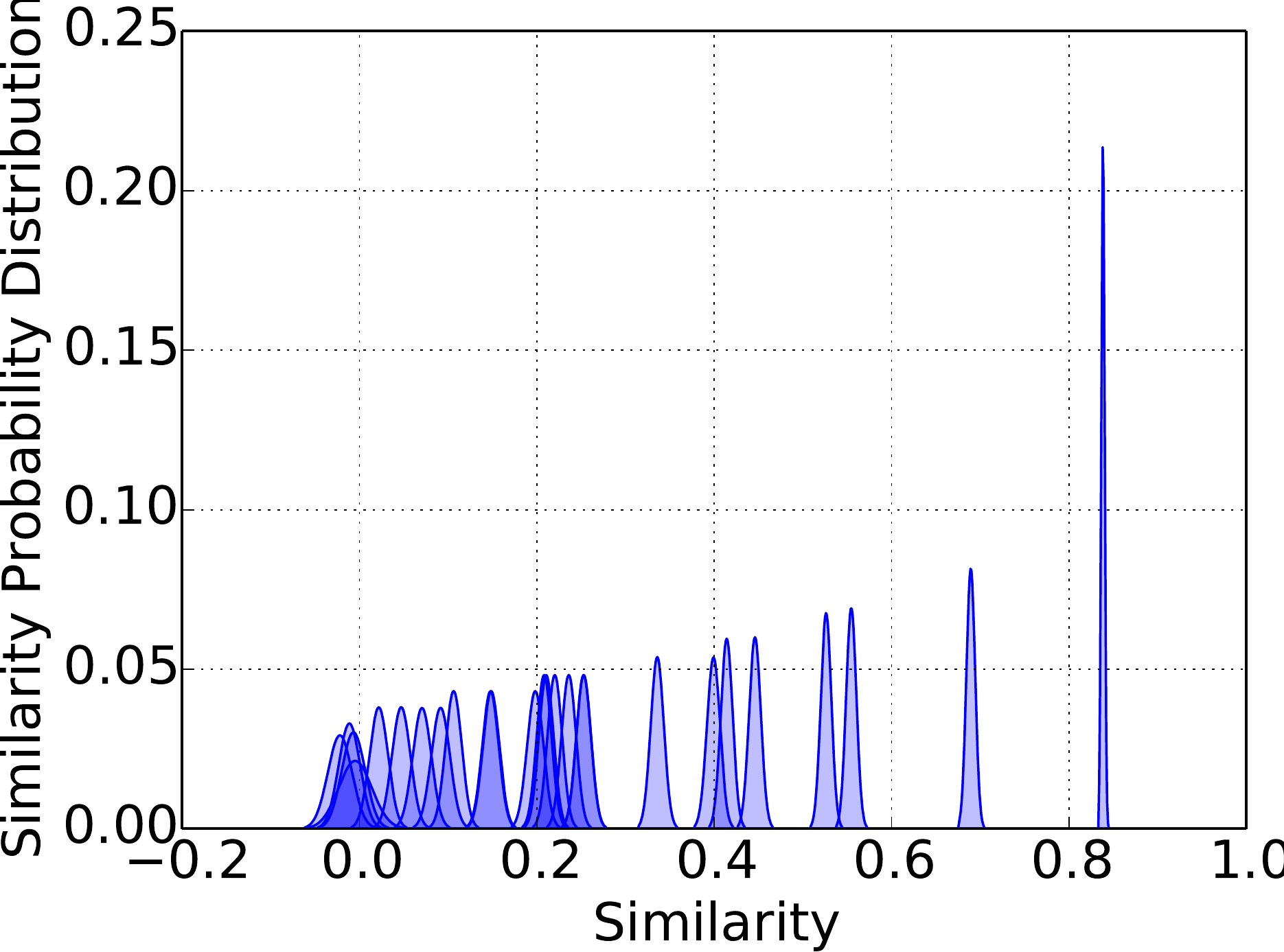}\label{figure:probabilitydist_sample_book}}
  \hfill
\centering
  \subfloat[]{\includegraphics[width=0.32\textwidth]{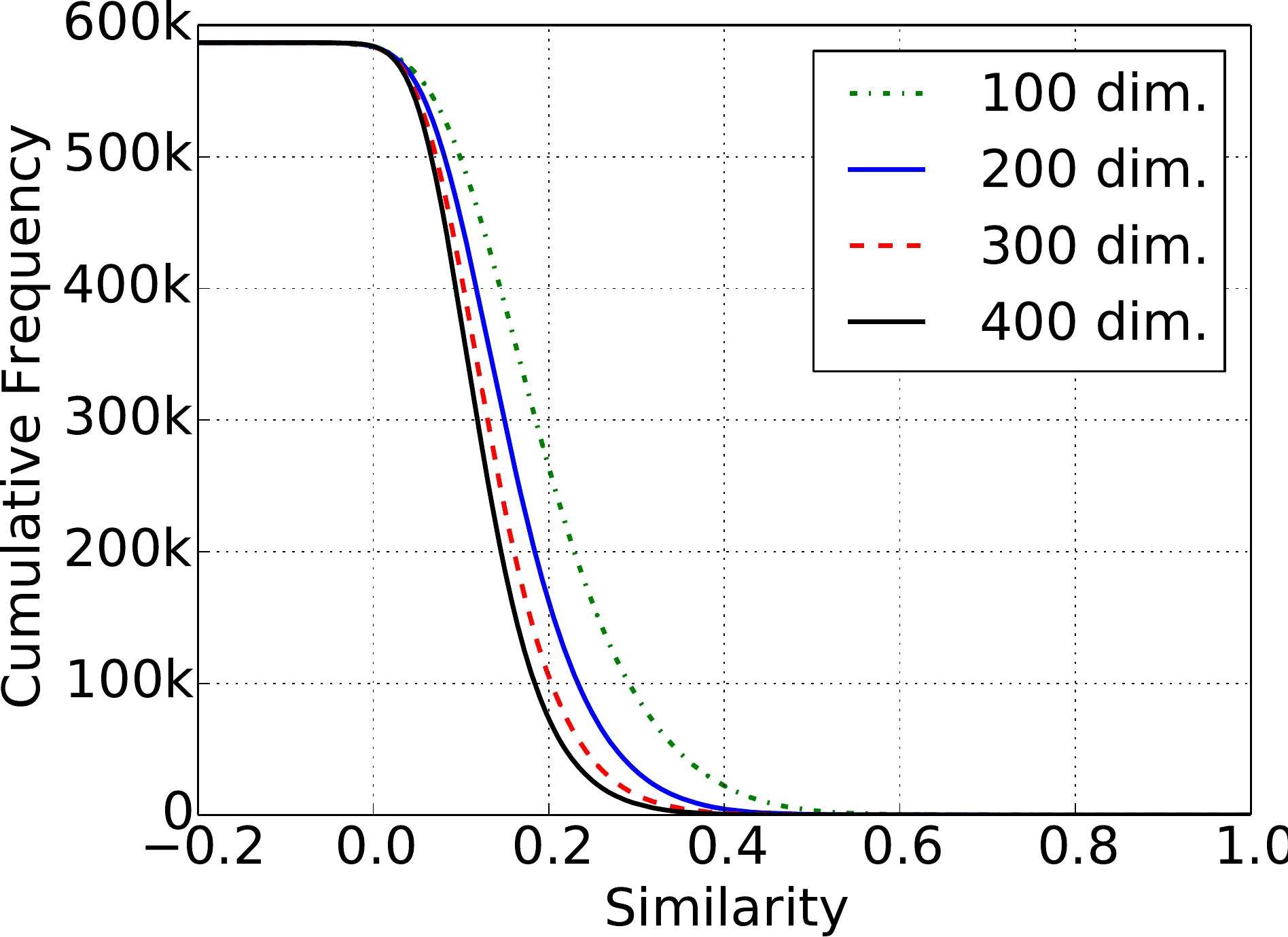}\label{figure:cdf_zoomedout}}
  \hfill
\centering
  \subfloat[]{\includegraphics[width=0.3\textwidth]{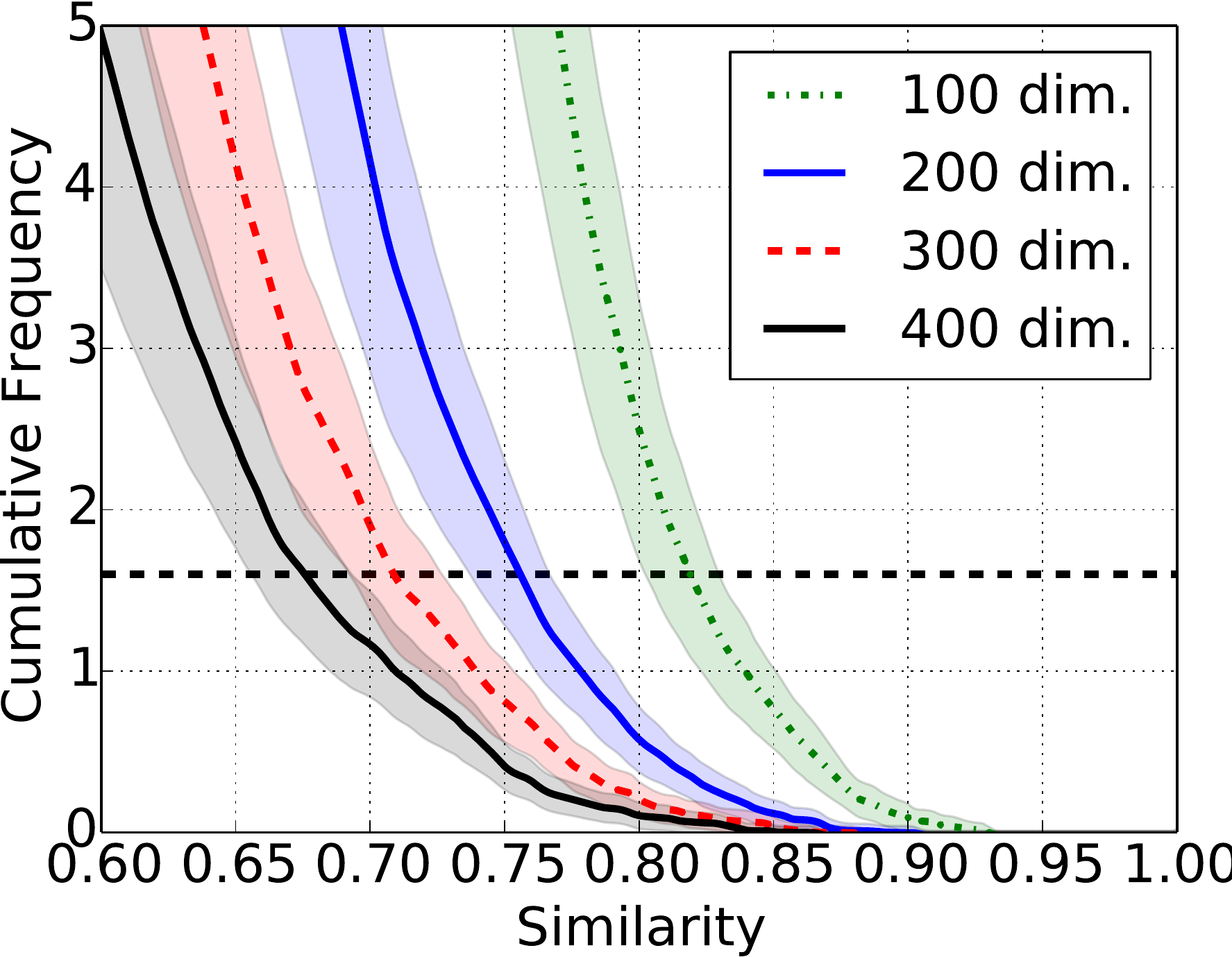}\label{figure:cdf_zoomed}}
  \hfill
  \vspace{-0.4cm}
  \caption{(a) Probability distribution of similarity values for the term \emph{Book} to some other terms (b) Mixture of cumulative probability distributions of similarities in different dimensions (c) Expected number of neighbours around an arbitrary term with confidence interval. The average number of synonyms in WordNet ($1.6$) is shown by the dash-line.}
  \label{figure:neighbours}
  \vspace{-0.4cm}
\end{figure*}
As seen, the different similarities of a pair of terms, achieved from different embedding models with the same training phases are slightly different. Intuitively, we assume that these similarity values follow a normal distribution such that we can consider every similarity value as a probability distribution, built based on the similarity values of the same pair in different models.

To estimate this probability distribution, for every dimension, we create five identical SGNS models following the setup in Section~\ref{sec:sec31}.  Figure~\ref{figure:probabilitydist_sample_book} shows the probability distribution of similarities for term \emph{Book} to 25  terms in different similarity ranges\footnote{we do not plot all to maintain the readability of the plot}. We observe that by decreasing the similarity, the variation of the probability distributions increases, reflecting the increase in uncertainty empirically observed between two models in the previous section.

We use these probability distributions to provide a representation of the expected number of neighbours around an arbitrary term in the spectrum of similarity values. For this purpose, we calculate the mixture of cumulative distribution functions of the probability distributions subtracted from 1, showed in Figure~\ref{figure:cdf_zoomedout}. The values on this plot indicate the number of expected neighbours in the area between the given similarity value to the term (similarity one). This representation of the expected number of neighbours in Figure~\ref{figure:cdf_zoomedout} has two main benefits: (1) the estimation is continuous, and (2) it considers the effect of uncertainty and considers all the models. 

As noted before, the notion of \emph{arbitrary} term is in fact an average over the 100 representative terms. However, this are just a sample of all terms in the vocabulary. Therefore, in calculating the representation of the expected number of neighbours, we also consider the confidence interval around the mean. This interval is shown in Figure~\ref{figure:cdf_zoomed}. Here, the representation is zoomed on the lower left corner of Figure~~\ref{figure:cdf_zoomedout}. The area around each plot shows the confidence interval of the estimation.

This continuous representation is used in the following for defining the threshold for the semantically highly related terms. 

\subsection{Similarity threshold}\label{sec:sec33}
Given the representation of the expected number of neighbours around the arbitrary term, the question is ``\emph{what is the best threshold for filtering the highly related terms?}''. This is of course a debatable question since the analytical approach attempts to measure the human understanding of synonymity. However, we hypothesise that since this general threshold tries to separates the highly related terms for an arbitrary term, it can be estimated from the average number of synonyms over the terms in language. Therefore, we transform the above question in a new question: ``\emph{What is the expected number of synonyms for a word in English?}'' 

To answer this, we exploit WordNet. We consider the distinct terms in the related synsets to a term as its synonyms, while putting out the multi word terms (e.g. Natural Language Processing, shown in WordNet by concatenating with underlines) since in creating the word embedding models we consider them as separated terms. The average number of synonyms over all the $147306$ terms of WordNet is $1.6$, while the standard deviation is $3.1$.

Using the mean value, we define our threshold for each dimensionality as the point where the estimated number of neighbours in Figure~\ref{figure:cdf_zoomed} is equal to 1.6. We also consider an upper and lower bound for this threshold based on the points that the confident intervals cross the approximated mean. The results are shown in Table~\ref{tbl:threshold_value}.
\begin{table}
\begin{center}
\caption{Potential thresholds}
\vspace{-0.4cm}
\begin{tabular}{c | c c c }
\multirow{2}{*}{Dimensionality}&\multicolumn{3}{c}{Threshold Boundaries}\\
&Lower&\textbf{Main}&Upper\\\hline
100&0.802&\textbf{0.818}&0.829\\
200&0.737&\textbf{0.756}&0.767\\
300&0.692&\textbf{0.708}&0.726\\
400&0.655&\textbf{0.675}&0.693\\
\end{tabular}
\vspace{-0.9cm}
\label{tbl:threshold_value} 
\end{center}
\end{table}

In the following sections, we validate the hypothesis by evaluating the performance of the introduced thresholds with an extensive set of IR experiments.

\vspace{-0.3cm}
\section{Experimental Methodology}
\label{sec:experimentsetup}
We test the effectiveness of the potential threshold in an Ad hoc retrieval task on IR test collections by evaluating the results of applying various thresholds to retrieve the related terms. 

Our relevance scoring approach is based on the \emph{language model}~\cite{Ponte:1998} method as a widely used and established method in IR that has shown competitive results in various domains. In particular, we use the \emph{translation language model}~\cite{Berger:1999} which includes the similarity of related terms into the basic model. 

In the following, first we explain the translation language model when combined with word embedding similarity and then describe the details of our experimental setup.

\subsection{Translation Language Model}
In the language model~\cite{Ponte:1998}, the score of a document $d$ with respect to a query $q$ is considered to be the probability of generating the query by a model $M_d$ estimated based on the document: 
\begin{equation}\label{eq:languagemodel}
score(q,d)=P(q|M_d)=\prod_{t_q\in q}P(t_q|M_d)
\end{equation}

Typically, the model is a multinomial distribution and the probability is computed with a maximum likelihood estimator, together with some form of smoothing. This smoothing, while not being part of the original idea, is in the practice of LM-based methods of paramount importance. However, this not being the focus of this study, we use Dirichlet smoothing~\cite{Zhai:2001}, as many others have done, successfully, before us (~\cite{zuccon2015integrating,Karimzadehgan:2010a,vulic2015monolingual}).

Berger and Lafferty~\shortcite{Berger:1999} introduced translation models as an extension to the language modelling. A translation model introduces in the estimation of  $P(q|M_d)$ a translation probability $P_T$, defined on the set of terms, always used in its conditional form $P_T(t|t')$ and interpreted as the probability of observing term $t$, having observed term $t'$. 
\begin{equation}\label{eq:translationmodel}
P(q|M_d)=\prod_{t_q\in q}\left(\sum_{t_d\in d}P_T(t_q|t_d)P(t_d|M_d)\right)
\end{equation}

The estimation of the model and specially the translation probability $P_T$ have been addressed by various approaches during the last two decades. Recently, Zuccon et al.~\cite{zuccon2015integrating} integrates word embedding into the  translation language model, showing potential improvement. In their work, they follow a $k$-NN approach to select the most similar terms for each query term in word embedding and estimate $P_T$ based on the similarity of the extended terms to the query term.

Similar to their work, we use the translation language model enhanced with word embedding and reproduce some of their experiments. However, instead of the $k$-NN approach, we apply our introduced thresholds (Section~\ref{sec:sec33}) to filter the similar terms.

\subsection{Experiments Setup}
We evaluate our approach on 5 test collections: combination of TREC 1 to 3, TREC-6, TREC-7, and TREC-8 of the AdHoc track, and TREC-2005 HARD track. Table~\ref{tbl:datasets} summarises the statistics of the test collections. For pre-processing, we apply the Porter stemmer and remove stop words using a small list of 127 common English terms. 
\begin{table}
\begin{center}
\caption{Test collections} 
\vspace{-0.3cm}
\begin{tabular}{l |  C{2.3cm} | c}
Name & Collection & \# Doc \\\hline
TREC 6 & Disc4\&5 & 551873\\
TREC 7, 8 & Disc4\&5 without CR & 523951 \\
HARD 2005& AQUAINT & 1033461 \\
\end{tabular}
\label{tbl:datasets} 
\end{center}
\vspace{-0.6cm}
\end{table}

In order to compare the performance of the potential thresholds, we test a variety of the threshold values in each dimension: for dimension 100, $\{$0.67, 0.70, 0.74, 0.79, 0.81, 0.86, 0.91, 0.94, 0.96$\}$, 200 dimension $\{$0.63, 0.68, 0.71, 0.73, 0.74, 0.76, 0.78, 0.82$\}$, 300 dimension, $\{$0.55, 0.60, 0.65, 0.68, 0.70, 0.71, 0.73, 0.75$\}$, and 400 dimension $\{$0.41, 0.54, 0.61, 0.64, 0.66, 0.68, 0.70, 0.71, 0.75$\}$. In addition to the threshold-based approach, we test the $k$-NN approach where $N$ is tested with $\{1, 2, 3, 5, 7, 10\}$ values.

We set the basic language model as baseline and test the statistical significance of the improvement of all the results with respect to it (indicated by the symbol $\dagger$). Since the parameter $\mu$ for Dirichlet smoothing of the translation language model is shared between the methods, the choice of parameters is not explored as part of this study. We select $\mu$ to 1000 as suggested in related studies. The statistical significance test are done using the two sided paired $t$-test and statistical significance is reported for $p<0.05$.

The evaluation of retrieval effectiveness is done with respect to MAP and NDCG@20, as standard measures. However, our initial experiments showed that using similar terms  retrieved a substantial proportion of unjudged documents. 
 Therefore, in order to provide a more fair evaluation framework, we consider MAP and NDCG over the condensed lists~\cite{Sakai:2007}\footnote{The condensed lists are used by adding the -J parameter to the trec\_eval command parameters}.

\vspace{-0.3cm}
\section{Results and Discussion}
\label{sec:evaluation}
\begin{figure*}[t!]
 \centering
  \subfloat[TREC-Adhoc-6]{\includegraphics[width=0.24\textwidth]{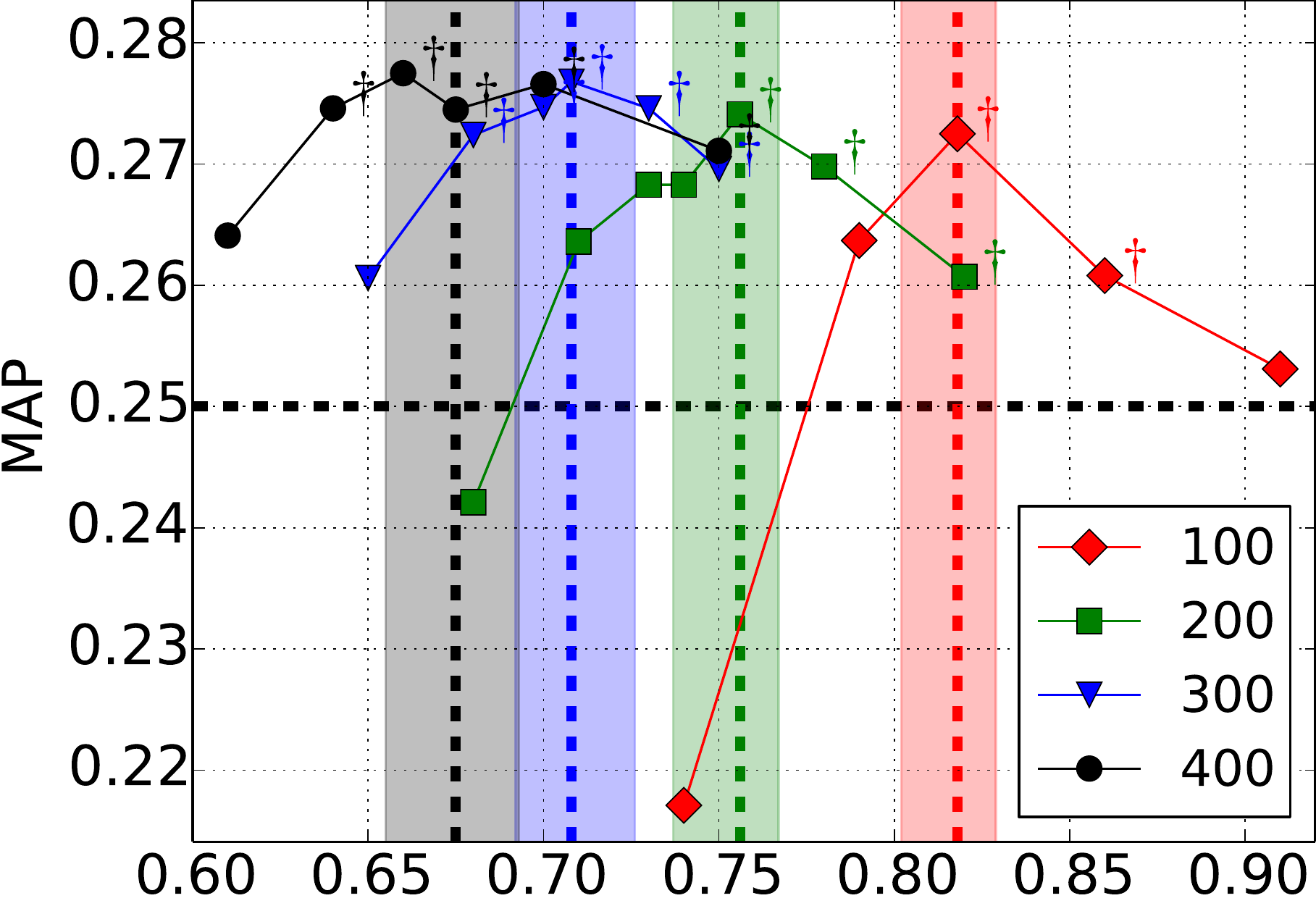}}
  \hfill
  \subfloat[TREC-Adhoc-7]{\includegraphics[width=0.24\textwidth]{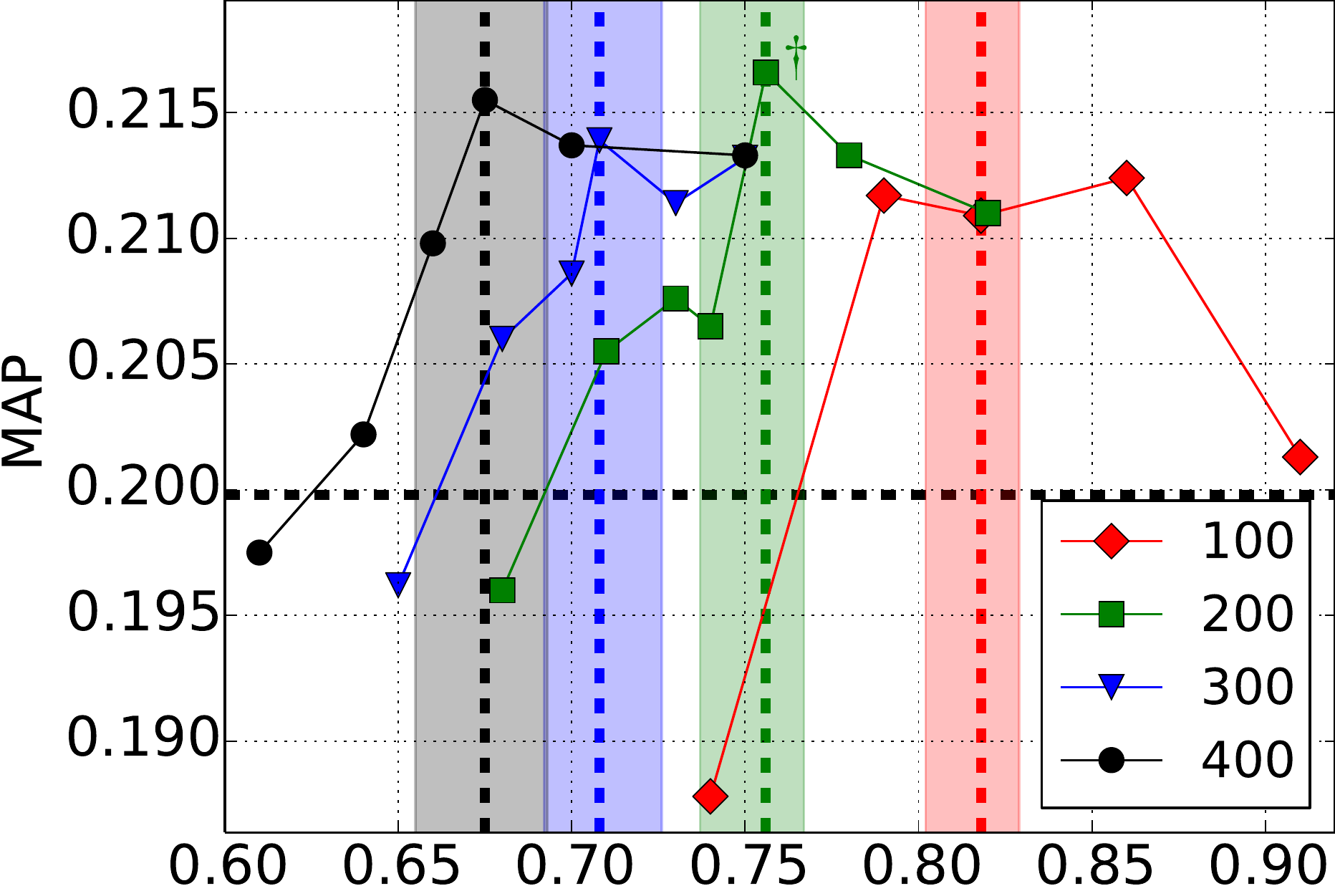}}
  \hfill
  \centering
  \subfloat[TREC-Adhoc-8]{\includegraphics[width=0.24\textwidth]{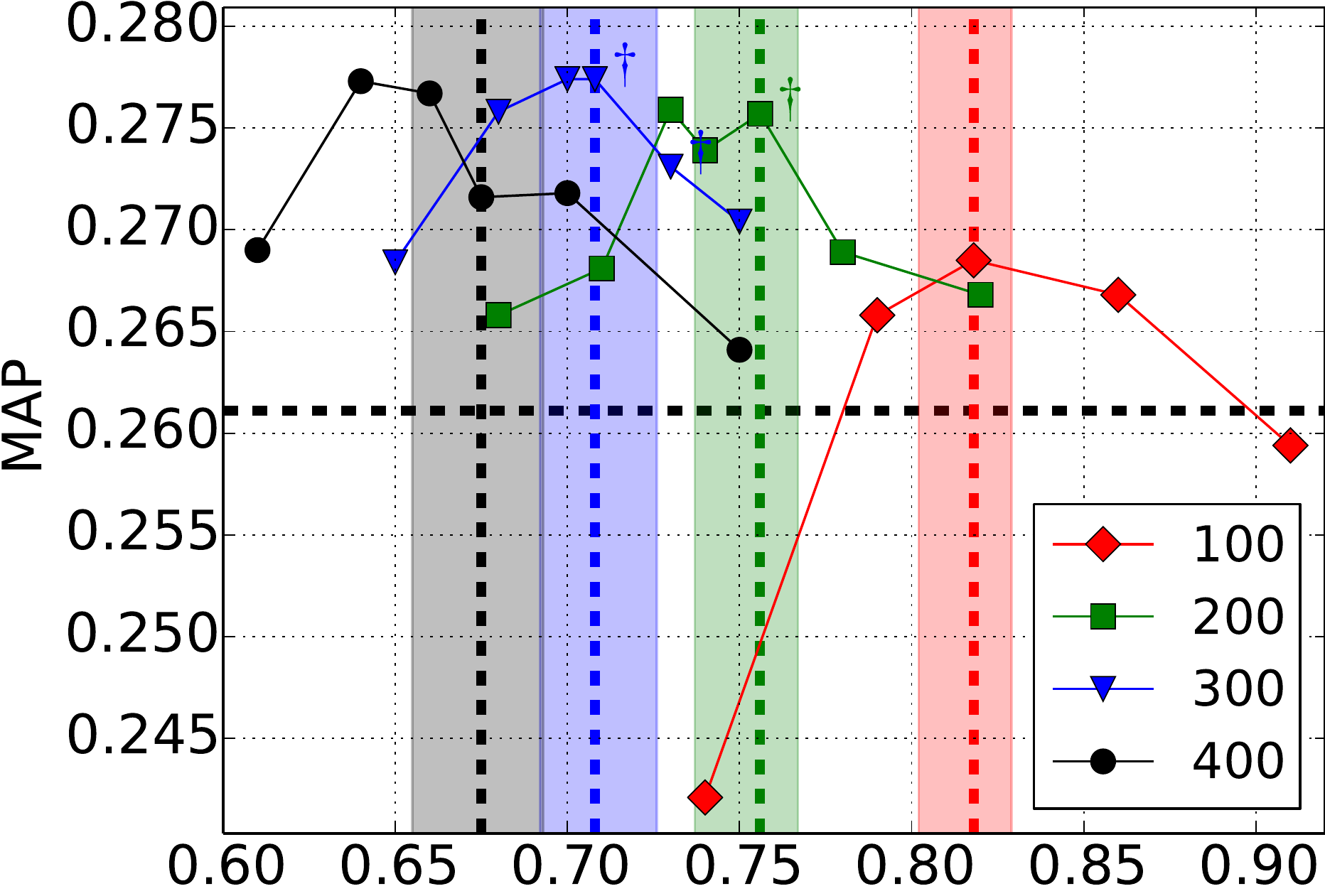}}
  \hfill
  \centering
  \subfloat[TREC-HARD-14]{\includegraphics[width=0.24\textwidth]{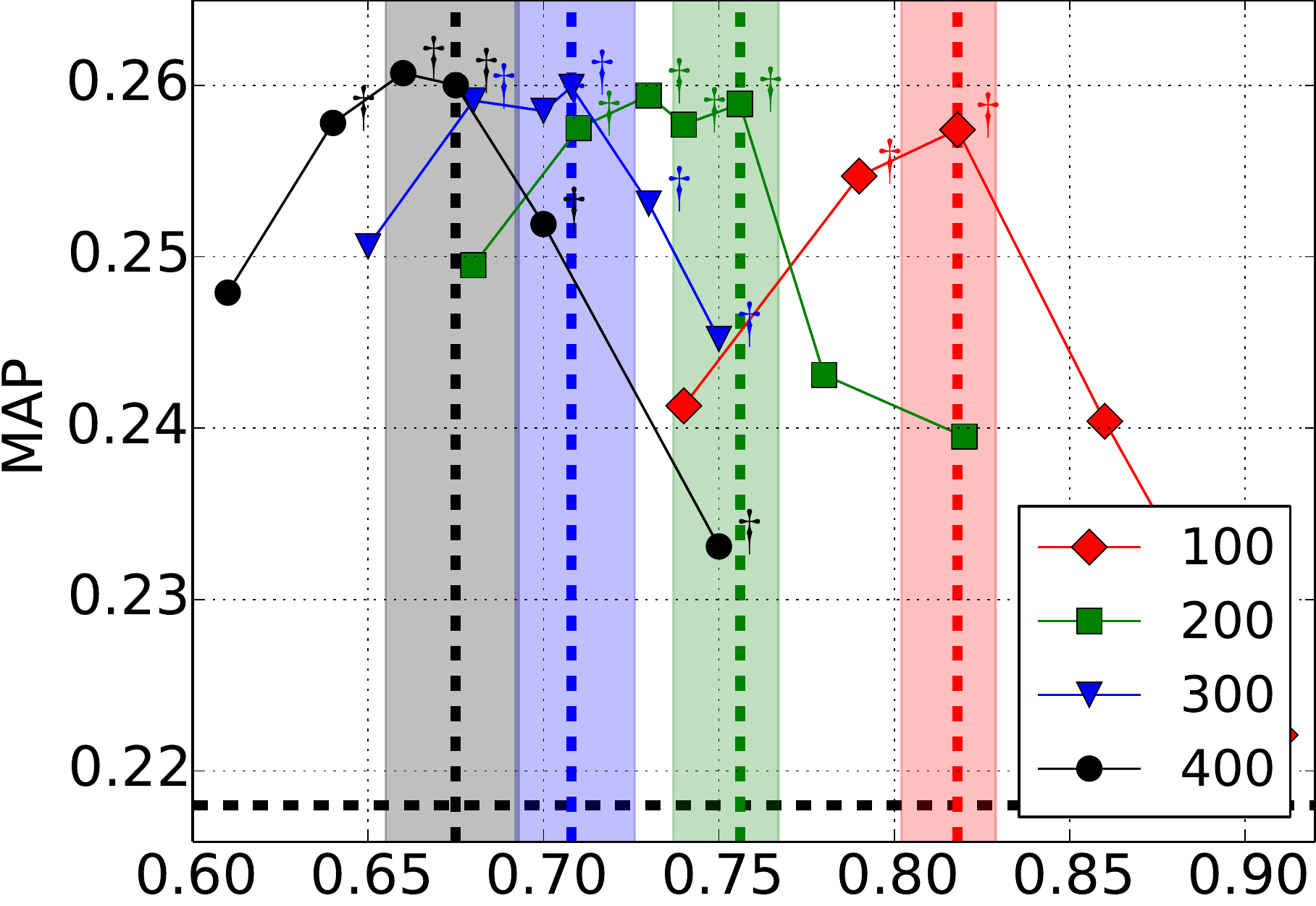}}
  \hfill 
  \centering
    \subfloat[TREC-Adhoc-6]{\includegraphics[width=0.24\textwidth]{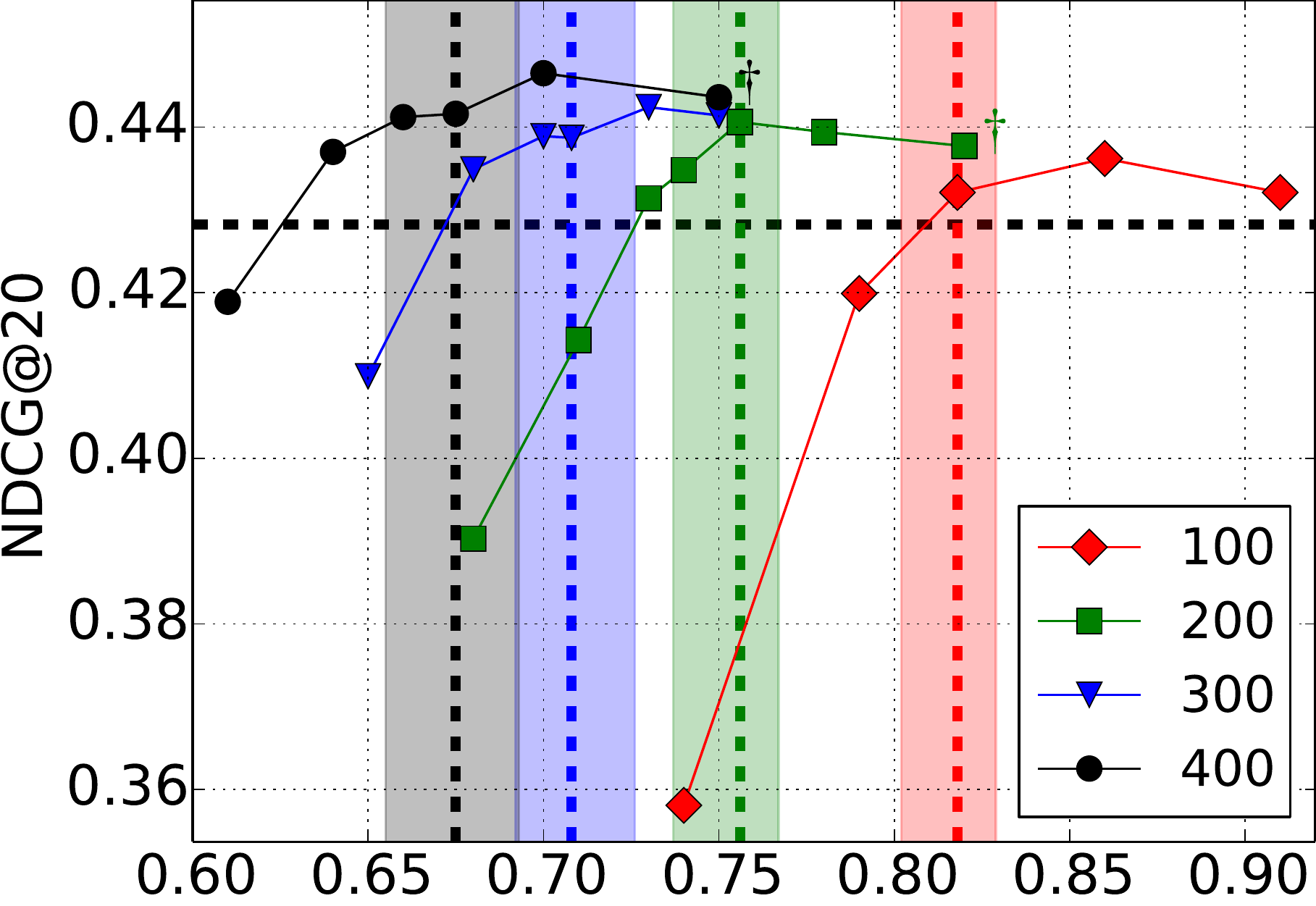}}
  \hfill
  \subfloat[TREC-Adhoc-7]{\includegraphics[width=0.24\textwidth]{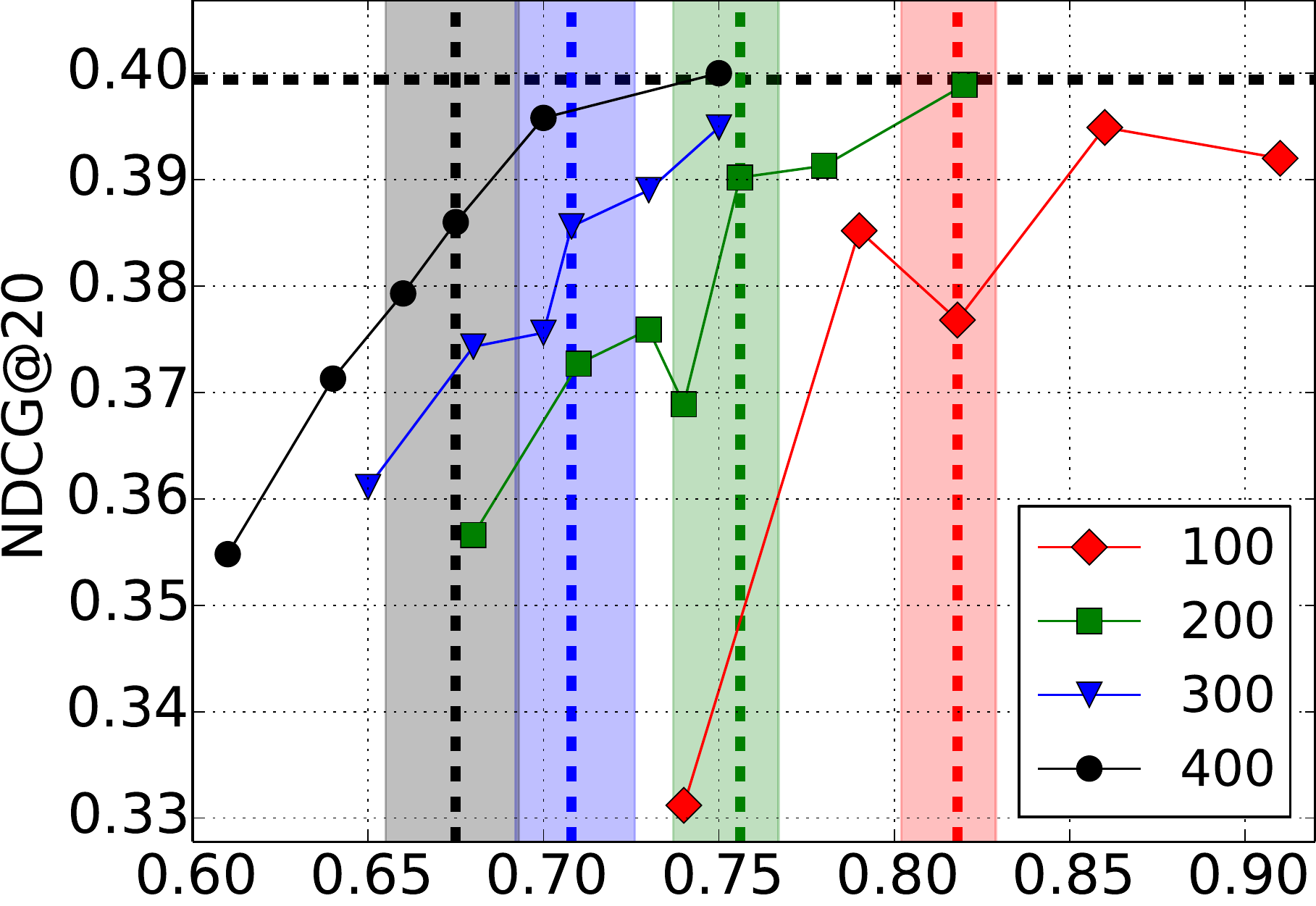}}
  \hfill
  \subfloat[TREC-Adhoc-8]{\includegraphics[width=0.24\textwidth]{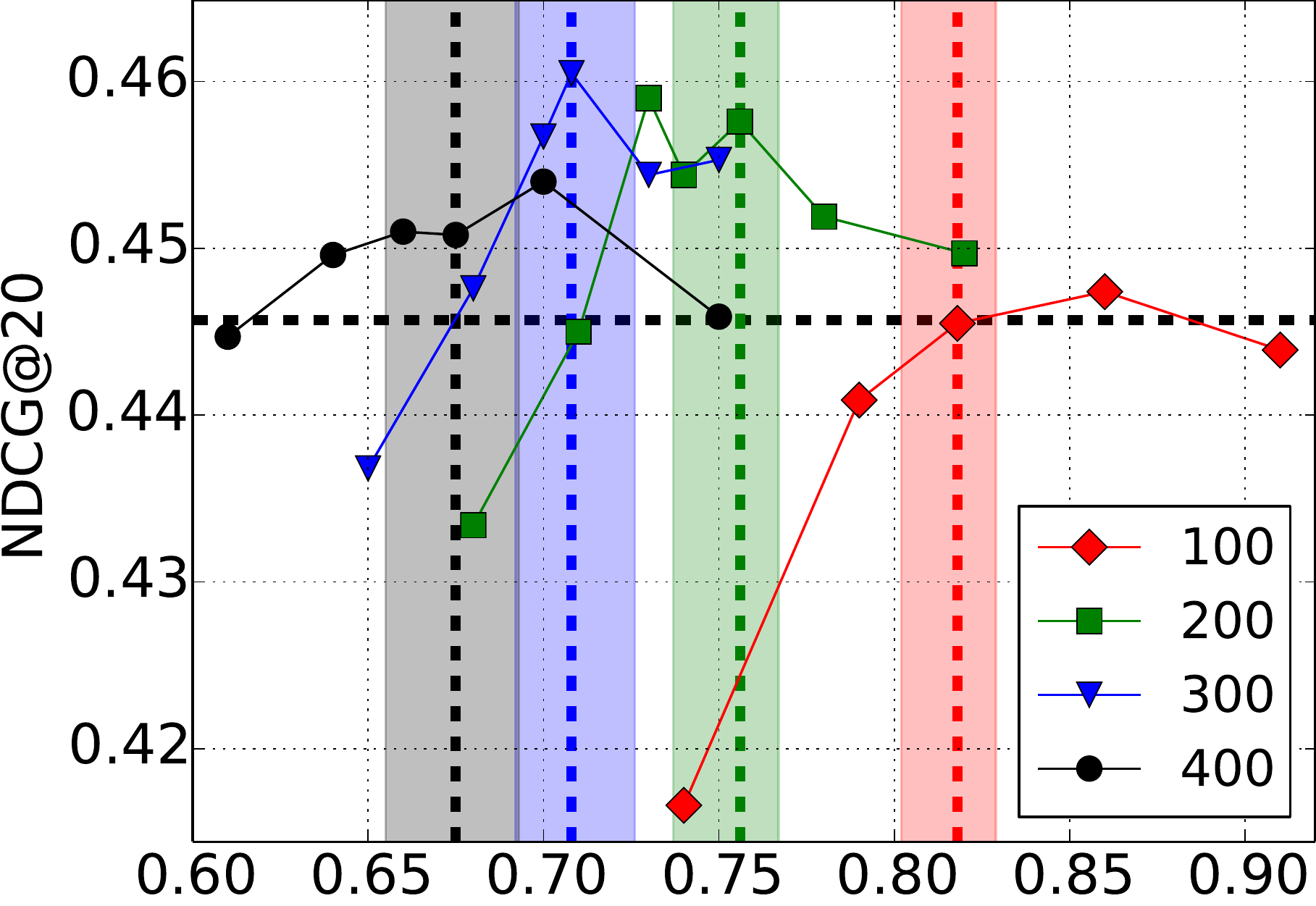}}
  \hfill
  \centering
  \subfloat[TREC-HARD-14]{\includegraphics[width=0.24\textwidth]{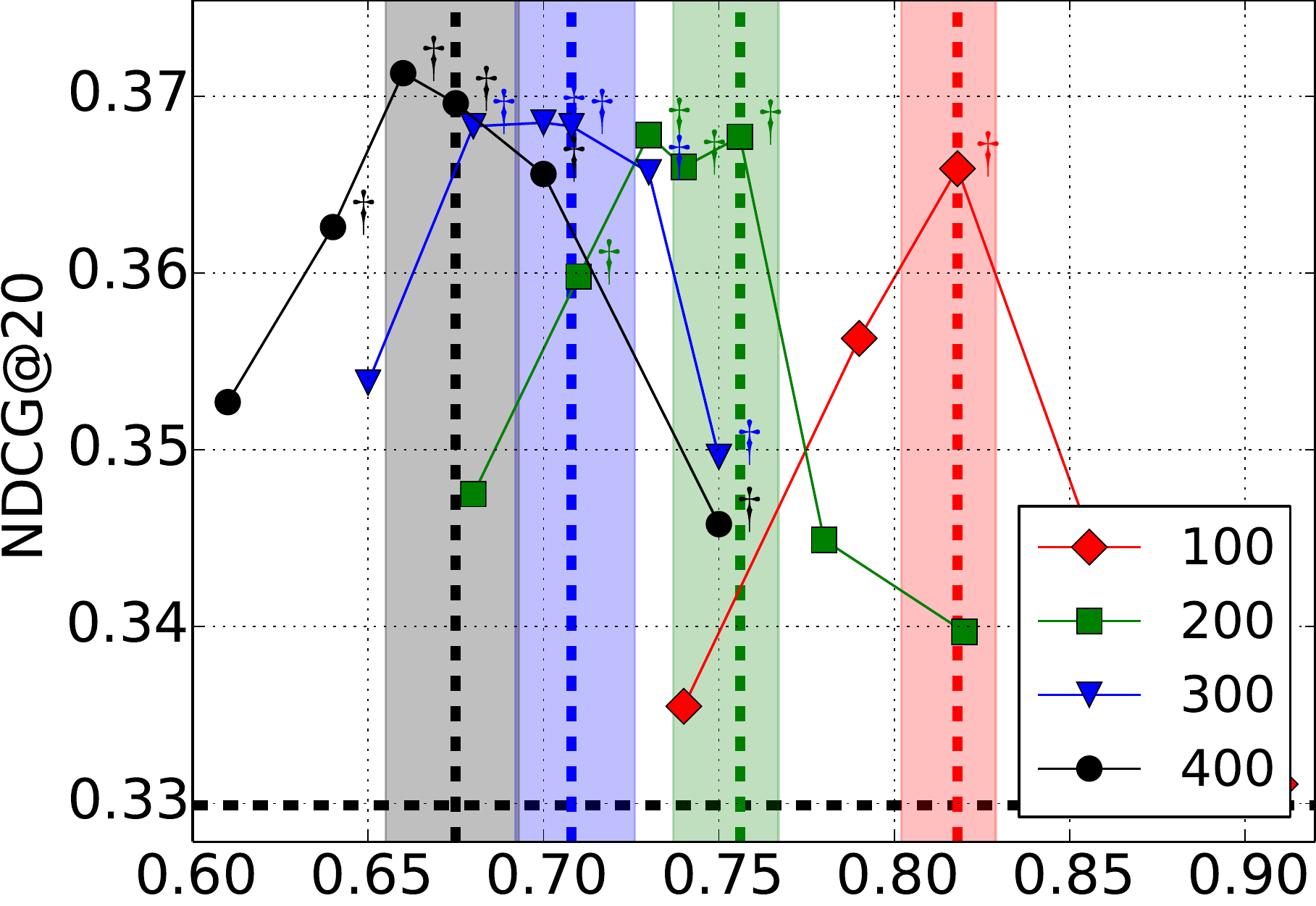}}
  \hfill
  \vspace{-0.3cm}
  \caption{MAP and NDCG@20 evaluation of the TREC-6, TREC-7, TREC-8 Adhoc, and TREC-2005 HARD for different thresholds and dimensions. Significance is shown by $\dagger$. Vertical lines indicate our thresholds in different dimensions. To maintain visibility, points with very low performance are not plotted.}
\vspace{-0.3cm}
  \label{figure:result}
\end{figure*}

\begin{table*}
\begin{center}
\caption{In each cell, the top value shows the result of the potential threshold and the bottom reports the optimal value (shown as - when equal to our threshold value). $\dag$ indicates a significant difference to the baseline. There is no significance difference between the results of the optimal value and our threshold.} 
 \vspace{-0.3cm}
\scriptsize
\begin{tabular}{c | L{1.0cm} L{1.0cm} | L{1.0cm} L{1.0cm} | L{1.0cm} L{1.0cm} |  L{1.0cm} L{1.0cm}}
\hline
\multirow{2}{*}{\textbf{Collection}} & \multicolumn{2}{c|}{\textbf{100 (0.81)}}
 &  \multicolumn{2}{c|}{\textbf{200 (0.74)}} & \multicolumn{2}{c|}{\textbf{300 (0.69)}} & \multicolumn{2}{c}{\textbf{400 (0.65)}}\\
& \textbf{MAP} & \textbf{NDCG} & \textbf{MAP} & \textbf{NDCG} & \textbf{MAP} & \textbf{NDCG} & \textbf{MAP} & \textbf{NDCG} \\\hline

\multirow{2}{*}{TREC-6}&\cellcolor{gray!20}0.273$\dagger$&\cellcolor{gray!20}0.432&\cellcolor{gray!20}0.274$\dagger$&\cellcolor{gray!20}0.441&\cellcolor{gray!20}0.277$\dagger$&\cellcolor{gray!20}0.439&\cellcolor{gray!20}0.275$\dagger$&\cellcolor{gray!20}0.442\\
&-&0.436&-&-&-&0.442&0.278$\dagger$&0.447\\\hline
\multirow{2}{*}{TREC-7}&\cellcolor{gray!20}0.211&\cellcolor{gray!20}0.377&\cellcolor{gray!20}0.217$\dagger$&\cellcolor{gray!20}0.390&\cellcolor{gray!20}0.214&\cellcolor{gray!20}0.386&\cellcolor{gray!20}0.215&\cellcolor{gray!20}0.386\\
&0.212&0.395&-&0.399&-&0.395&-&0.400\\\hline
\multirow{2}{*}{TREC-8}&\cellcolor{gray!20}0.269&\cellcolor{gray!20}0.446&\cellcolor{gray!20}0.276$\dagger$&\cellcolor{gray!20}0.458&\cellcolor{gray!20}0.277$\dagger$&\cellcolor{gray!20}0.461&\cellcolor{gray!20}0.272&\cellcolor{gray!20}0.451\\
&-&0.447&-&0.459&-&-&0.277&0.454\\\hline
\multirow{2}{*}{HARD}&\cellcolor{gray!20}0.257$\dagger$&\cellcolor{gray!20}0.366$\dagger$&\cellcolor{gray!20}0.259$\dagger$&\cellcolor{gray!20}0.368$\dagger$&\cellcolor{gray!20}0.260$\dagger$&\cellcolor{gray!20}0.368$\dagger$&\cellcolor{gray!20}0.260$\dagger$&\cellcolor{gray!20}0.370$\dagger$\\
&-&-&-&-&-&-&0.261$\dagger$&0.371$\dagger$\\\hline

\end{tabular}
\label{tbl:eval} 
\vspace{-0.3cm}
\end{center}
\end{table*}

The evaluation results of the MAP and NDCG@20 measures on the 4 test collections, with vectors in 100, 200, 300, and 400 dimensions are shown in Figure~\ref{figure:result}.  For each dimension our threshold and its confidence interval are shown with vertical lines. Significant differences of the results to the baseline are marked on the plots using the $\dagger$ symbol. Table~\ref{tbl:eval} summarizes the results of the optimal as well as potential thresholds. 

Based on the results, we gain significantly better performance in all the collections at least in one of the threshold values. Except for TREC-7, we observe similar results with both the evaluations measures. 

The plots show that the performance of the method is highly dependent on the choice of the threshold value. In general, we can see a trend in all  dimensions: the results tend to improve till reaching a peak (optimal threshold) and then decrease and finally converge to the baseline. Based on this general behaviour, we can assume that including the terms before the optimal threshold introduces noise and deteriorates the results while after it, the terms are filtered too strictly and there are still related terms to improve the results. Comparing the results of the optimal and potential threshold, in most the cases the optimal one is either the same or in the confidence area of our introduced threshold such that there is no statistically  significant difference between the optimal and our threshold.

\begin{figure*}
  \centering
 \subfloat[]{\includegraphics[width=0.3\textwidth]{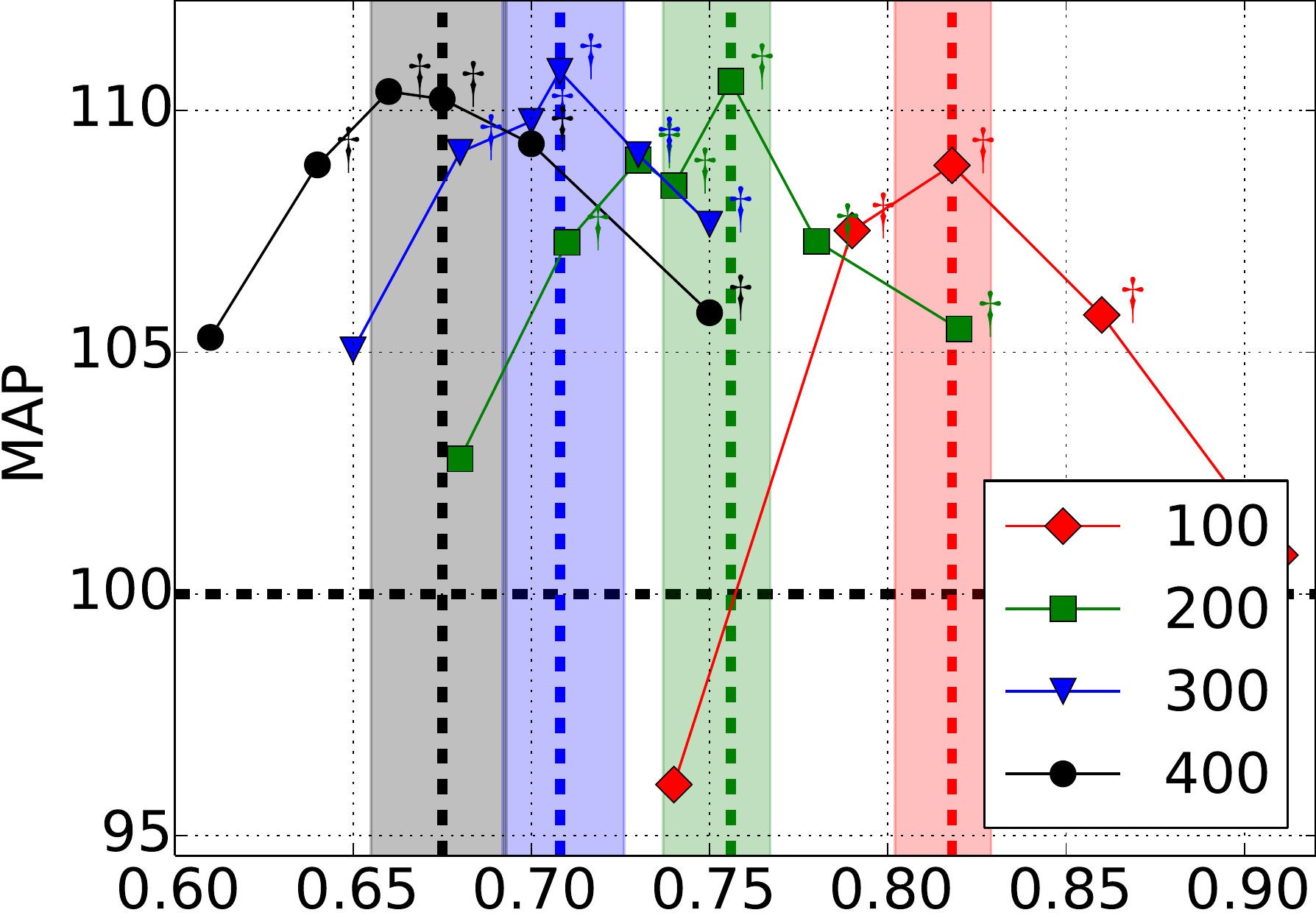}\label{figure:result_all_map_thr}}
  \hfill
  \centering
\subfloat[]{\includegraphics[width=0.3\textwidth]{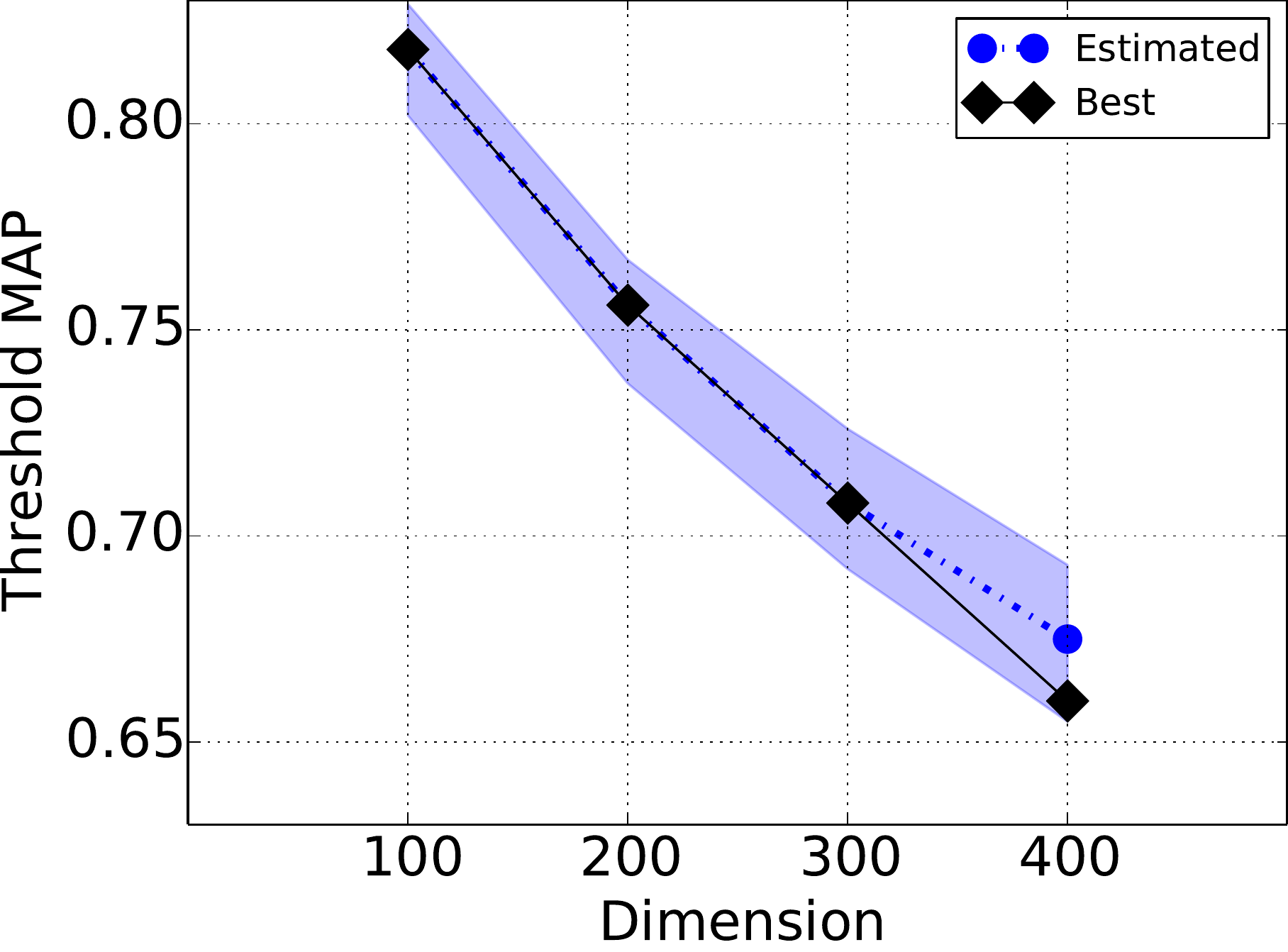}\label{figure:result_bestvsest_J_map}}
  \hfill
   \centering
   \subfloat[]{\includegraphics[width=0.3\textwidth]{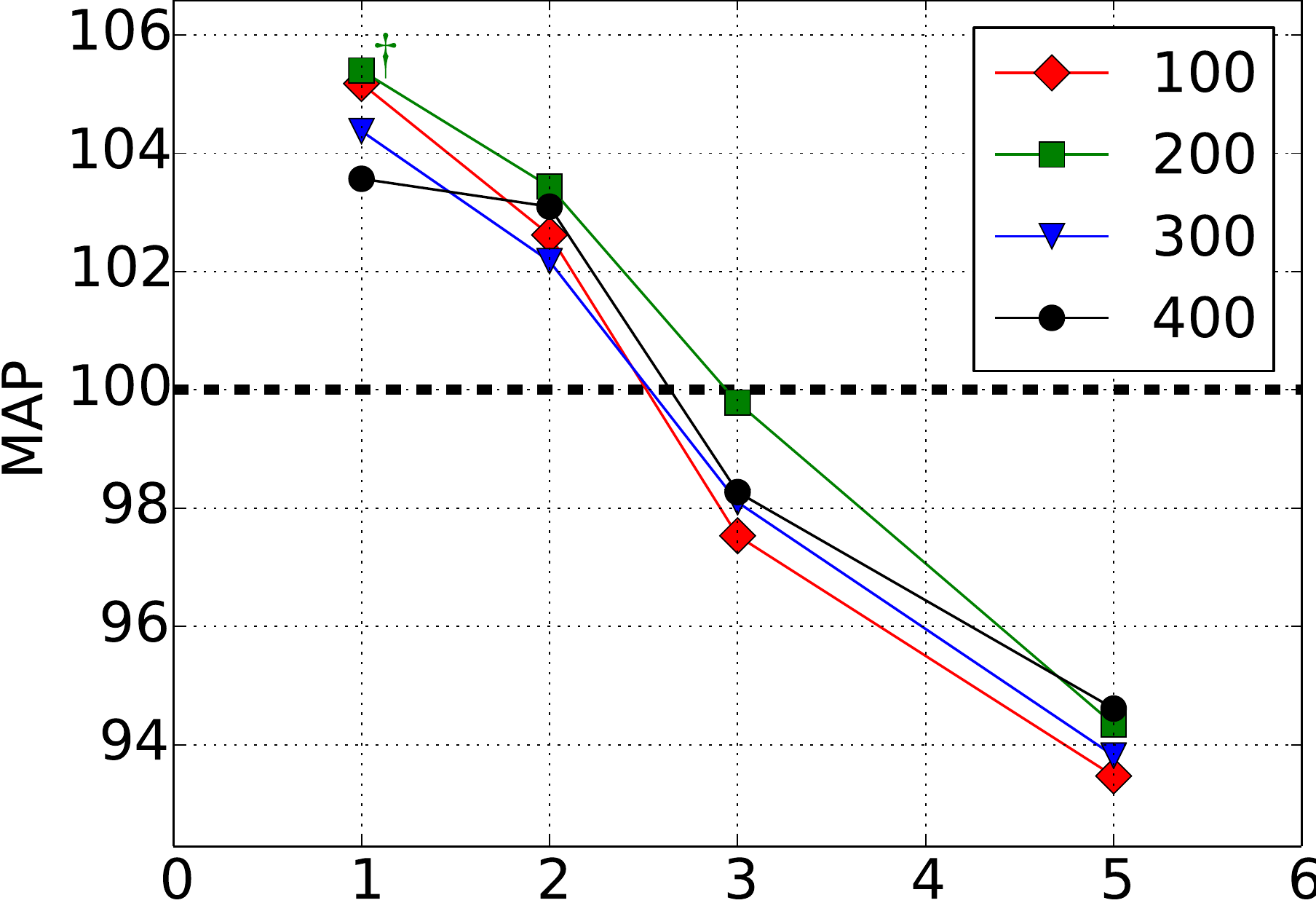}\label{figure:result_all_J_map_STD_CNT}}
\hfill
\vspace{-0.5cm}
  \caption{All the results in MAP measure: (a,c) Improvement of the models with respect to the original language model (baseline), aggregated over all the collections (b) The potential and optimal thresholds in different dimensions where results are aggregated over all the collections. (c) same as Figure a but using $k$-NN approach.}
\vspace{-0.3cm}
  \label{figure:result_all}
\end{figure*}

In order to have an overview on all the models, we calculate the gain of each model over the baseline and averaged the gains on the five collections. The results for MAP\footnote{The NDCG results are very similar and not shown for space}  are depicted in Figures~\ref{figure:result_all_map_thr}. Also the potential threshold and its confidence interval are compared with the optimal one in different dimensions in Figure~\ref{figure:result_bestvsest_J_map}. 
Our threshold is optimal for dimensions 100, 200, and 300, and in dimension 400 it is statistically indistinguishable from the optimal. This results justifies the choice of the introduced threshold as a generally stable and effective cutting-point for identifying highly related terms.

For completeness, we also conducted experiments on the $k$-NN approach. The results in Figure~\ref{figure:result_all_J_map_STD_CNT} show the very weak performance of the $k$-NN approach for MAP measure such that it has slightly better than baseline for $k$ equal to 1 and 2 and then radically deteriorates by increasing $k$. 
\begin{table}
\scriptsize
\begin{center}
\vspace{-0.3cm}
\caption{Examples of similar terms, selected with the potential threshold}
\begin{tabular}{L{7.5cm}}\hline
book: publish, republish, foreword, reprint, essay\\
eagerness: hoping, anxious, eagerness, willing,wanting\\
novel: fiction, novelist, novellas, trilogy\\
microbiologist: biochemist, bacteriologist, virologist\\
shame: ashamed\\
guilt: remorse\\
Einstein: relativity\\
estimate, dwarfish, antagonize: no neighbours\\\hline
\end{tabular}
\label{tbl:threshold_example} 
\vspace{-0.5cm}
\end{center}
\end{table}

To understand this behaviour let us take a closer look at the selected terms. Table~\ref{tbl:threshold_example} shows some examples of the retrieved terms when using the word embedding model with 300 dimension with the our threshold (same as optimal in this dimension). The examples show the strong differences in the number of similar words for various terms. The mean and standard deviation of the number of similar terms for the $508$ query terms of the tasks is $1.5$ and $3.0$ respectively. Almost half of the terms are not expanded at all. An interesting observation is the similarity between this calculated mean and standard deviation  and the aggregated number of synonyms we observed in WordNet in Section~\ref{sec:sec33}---mean of $1.6$ and standard deviation of $3.1$. It appears that although the two semantic resources cast the notion of similarity in very different ways and their provided sets of similar terms are very different, they correspond to very similar distribution of the number of related terms.

\vspace{-0.3cm}
\section{Conclusion and Future Work}
\label{sec:conclusion}
We have analytically explored the thresholds on similarity values of word embedding to select related terms. This threshold is estimated based on a novel representation of the neighbours around an arbitrary term which is  continuous and  benefits from addressing the issue of uncertainty in similarity values of modern word embedding models.

We extensively evaluate the application of the suggested threshold on four information retrieval collections. The results show superior performance when using our threshold such that its results are either equal to or statistically indistinguishable from the optimal results, achieved by extensive search on the parameter space. 

\vspace{-0.3cm}
\bibliographystyle{abbrv}
\bibliography{refer}  

\end{document}